\documentclass{article}
\usepackage[utf8]{inputenc}

\usepackage{amsmath}

\usepackage{cite}
\usepackage{amsmath,amssymb,amsfonts}
\usepackage{algorithmic}
\usepackage{geometry}
\usepackage{graphicx}
\usepackage{textcomp}
\usepackage{xcolor}
\usepackage{amsthm}

\def\BibTeX{{\rm B\kern-.05em{\sc i\kern-.025em b}\kern-.08em
    T\kern-.1667em\lower.7ex\hbox{E}\kern-.125emX}}
\begin{document}

\title{BUDS: Balancing Utility and Differential Privacy by Shuffling}

\author{
\begin{tabular}{*3c}
   \large Poushali Sengupta$^1$  & \large Sudipta Paul$^{2,3}$  & \large Subhankar Mishra$^{2,3}$ \\
    &&\\
    \multicolumn{3}{c}{\textit{${^1}$Department of Statistics, University of Kalyani}}\\
    \multicolumn{3}{c}{Kalyani, West Bengal, India - 741235}\\
    &&\\
    \multicolumn{3}{c}{\textit{${^2}$School of Computer Sciences,}}\\
    \multicolumn{3}{c}{\textit{National Institute of Science, Education and Research Bhubaneswar}}\\
    \multicolumn{3}{c}{India - 752050}\\
    &&\\
    \multicolumn{3}{c}{\textit{${^3}$Homi Bhaba National Institute, Anushaktinagar, Mumbai}}\\
     \multicolumn{3}{c}{India - 400094}\\
     &&\\
     \multicolumn{3}{c}{${^1}$tua.poushalisengupta@gmail.com}\\
     \multicolumn{3}{c}{$^{2,3}$\{sudiptapaulvixx, smishra\}@niser.ac.in}\\
     \end{tabular}
}


\maketitle

\begin{abstract}
Balancing utility and differential privacy by shuffling or \textit{BUDS} is an approach towards crowd sourced, statistical databases, with strong privacy and utility balance using differential privacy theory. Here, a novel algorithm is proposed using one-hot encoding and iterative shuffling with the loss estimation and risk minimization techniques, to balance both the utility and privacy. In this work, after collecting one-hot encoded data from different sources and clients, a step of novel attribute shuffling technique using iterative shuffling (based on the query asked by the analyst) and loss estimation with an updation function and risk minimization produces a utility and privacy balanced differential private report. During empirical test of balanced utility and privacy, BUDS produces $\epsilon = 0.02$ which is a very promising result. Our algorithm maintains a privacy bound of $\epsilon = ln [t/((n_1 - 1)^S)]$ and loss bound of $c' \bigg|e^{ln[t/((n_1 - 1)^S)]} - 1\bigg|$.  
\end{abstract}


\section{Introduction}
Crowd sourcing is a very powerful tool for collecting data, for example during recent pandemic like Covid-19 all the datasets have been collected from different hospitals, patients through this process only, for further researches to fight the situation. This method has been widely used by the software companies, the health service providers or the researchers in different fields which helps to collect the necessary data easily from the willing participants. But, maintaining the privacy of the participants in case of a survey concerning a sensitive issue or an attack strong enough to reveal the individual identity, has been a real challenge for a while now. Differential privacy (DP) provides a provable solution to this privacy concern of individual in a database with some trade offs.     
\par Differential privacy theory acknowledges the privacy concern of both the data analyst and the participants in a survey. DP promises that the statistically distributed answer to the question asked by the analyst will be in the neighbourhood of the actual answer but will not reveal the accurate one. This promise provides enough noise in the answer such that the individual's information can not be leaked from the aforementioned survey databases. Also, the analyst gets enough insight of the databases for further analysis without being able to extract any individual's information. 
\par A data analyst asks the queries through a query function to the database. The response to that query, over the entire database, is then generated using an algorithm based on the differential privacy theory (DP algorithm) in the form of a report. Here utility can be defined as, how much insight about that database can be extracted from the report. To maintain individual's privacy, an 100\% utility is not desirable at all. Therefore maintaining a proper balance between utility and privacy is an essential characteristics for a good and feasible DP algorithm irrespective of its two broad kinds, i.e. local DP(LDP) and centralized DP(CDP). RAPPOR\cite{b2}, PROCHLO (implementation of ESA)\cite{b3}, amplification by shuffling\cite{b4}, ARA\cite{b5} are some of the standard works accounting both the kinds LDP and CDP. But every work that is mentioned above does not have that optimal balance of utility and privacy. Hence to acknowledge this problem a balanced algorithm with better utility and privacy is proposed here.
\par The contributions of this work are following:
\begin{itemize}
    \item The introduction of the optimal condition for the privacy parameter $\epsilon$. It has been shown that this algorithm maintains the privacy of $\epsilon = ln [t/((n_1 - 1)^S)]$ , where, $\epsilon$ is the privacy parameter, $n_1$ is the first batch size, $S (S>1)$ is the number of shufflers and $t$ is the number of batches.
    \item The introduction of loss function updation to guarantee the longitudinal privacy. 
    \item  The introduction of risk function for balancing the differential trade-offs and choosing the optimal randomization scheme.
    
\end{itemize}
\par The paper is arranged in the following manners: following an introduction the necessary background knowledge is given in the section \ref{rw}, a utility and privacy balanced DP algorithm is proposed in the section \ref{architecture} to acknowledge the aforementioned problem, the proofs of utility and privacy for the proposed algorithm is given consecutively in the sections \ref{privacy} and \ref{utility}.

\section{Related Work}\label{rw}
The main concern of statistical database of sensitive information has always been to protect the "privacy" of "individual" from the attack of intruders. As a starting point, Dalenius desideratum \cite{b7} gives a hard bound on this privacy situation. the desideratum is following:\\
\textit{“Nothing about an individual should be learn-able from the database that could not be learned without access to the database!”}
\par By introducing "Differential privacy"(DP) in 2006 Dwork showed that \textit{Dalenius desideratum} is too strong to be useful in practice. The fundamental definitions, theorems and discussions to understand the concept of differential privacy is given in \cite{b1}. The main motto of DP is not to protect the entire database but to protect the sensitive data of individual of a database from the attackers. One of the main DP approaches LDP \cite{b8} is based on randomized technique where the noise get injected inside the data at the time of collection. Another approach CDP is a centralized approach where a trusted curator maintain the DP promises to the whole forest of data in a database.

\par The present DP models can be classified into these two broad categories. \textit{RAPPOR}\cite{b2} is currently the fastest implementation in LDP by using hash functions in a novel manner by using bloom filters, to produce noisy reports for a particular query. But the main drawback of RAPPOR is that the mechanism fails to provide enough utility. In 2017,  \textit{PROCHLO}\cite{b3} is introduced, that uses the \textit{Encode, Shuffle, Analyze} framework to provide CDP on the data base.\textit{ Amplification by Shuffling}\cite{b4} is another work on CDP in 2018 where the authors introduced a mechanism with \textit{oblivious shuffling} that provides $\epsilon$-LDP which satisfies $\mathcal{O}(\epsilon \sqrt{\log{\frac{1}{\delta}/ n}, \delta})$ centralized differential privacy(CDP) guarantee. The disadvantage of this work is the inability to guarantee the longitudinal privacy. In their work in 2020, \textit{ARA}\cite{b5} the authors describe a mechanism that provides CDP on \textit{RAPPOR} reports. Due to excessive noisy data, the main downside of this work is that the level of  utility is not more than $52.28\%$ in average, even though it is one of the fastest implementation in CDP. To recover these aforementioned problems in 2020,\cite{b3} the work was revisited and a new approach \cite{b6} was proposed that introduced report fragmentation to generate the final report to keep strong privacy promises for longitudinal database with maximised utility guarantee. But this framework involves a complex process and harder to implement. To simplify the architecture, the authors of this paper proposed the method of iterative shuffling that provides a significant balance between utility and privacy. This mechanism works with minimum noise that gives an impressive utility and privacy guarantee on the longitudinal data at the same time. This work\cite{b11} shows that the shuffle protocols for the widely studied selection problem requires exponentially higher sample complexity than the central model protocols which is a direct inspiration to this work. It is a special case of ESA framework. One of the direct inspirations was Li et al.'s paper\cite{b12}. The main motto of this paper\cite{b12} is to balance the trade off between utility and Privacy and it shows that the numerical nature of domain report is always a advantage for balancing these two. It introduces square wave (SW)  mechanism and expectation maximization with smoothing (EMS) mechanism where SW  exploits the numerical nature of the domain reports and EMS is applied to aggregate the histogram generated from SW mechanism to estimate the original distribution. The main drawback of this paper is the entire work is on LDP. Another work by Lecuyer et al.\cite{b10} presents the first certified  defence that is biased on a novel connection between robustness against adversarial example and DP, which not only works for large network and dataset , but also can be applied to the arbitrary model setup. Another work that inspires us\cite{b13}, introduces the capacity bounded differential privacy where the adversary that distinguishes the output distribution is assumed to be  not bounded in computational power. But this work does not explore the generalization in high probability and adaptivity in statistical generalization. This work has a future promise to explore into the informational geometry pf adversarial divergence. Another work on LDP\cite{b14} provides more accurate location privacy recommendation using matrics factorization. In addition they found out the best granularity of a location privacy preference and built a suitable method that can predict the location privacy preference accurately which inspires us too. But it was, solely LDP based with less utility in comparison to the normal matrix factorization technique and has a promise to go into more granularity depth with respect to time. 
\section{Architecture}\label{architecture}
In this work, after the collection of data from various sources with one hot encoding, a query function is applied on the data which the attributes that are required for generating reports and then the reported attributes tie up together to represents a single attribute and all attributes are transmitted to the shufflers for secret iterative shuffling. After this, finally the report is generated. This mechanism is named as $BUDS \ : \ Balancing \ Utility \ and \ Differential \ Privacy \ by \ Shuffling$. The whole process is divided into the following steps (figure \ref{architecture}):
\subsection{Collecting Data with One-Hot Encoding}
The data records are collected from various sources with one-hot encoding. Let, for a data record $X$, the domain of the x is $D$ which is a dictionary of elements and we want to estimate the frequency of the elements that is contained by the record $X$ in the domain $D$. When the cardinality of $D$ is not too large, the best way to encode data is one-hot encoding that is a group of bits among which the legal combinations of values are only those with a single high bit (1) and all other bits are low (0).  Now, the choice of encoding has a great impact on utility. When, the cardinality of $D$ is too large, one may use the sketching algorithm for encoding , but in practical it will not give a good utility as it is shown in the ESA revisited\cite{b6}. In this architecture, the preference is one-hot encoding instead of any other encoding to produce a result with minimum noise  that  always provides maximum utility while keeping $\epsilon$-Differential Privacy.  
\subsection{Shuffling}
A randomised mechanism is applied with shuffling which occurs repeatedly to the given data for producing randomised report to a particular query. This procedure is decomposed into two parts, in the first part the query function is applied to the data to get the important attributes for generating the final report and in the second part, all rows of each attributes are shuffled by the $S \ (S > 1)$ number of shuffler. 
\par
\textbf{Example 1: }If  a database contains names, ages, heights and weights of individuals where $name, age, \:height\:$ and$\: weight$ are four attributes and the query is \textit{"How many people are in database are less than 40 years old and have name starting with letter $'N'$?"}, then the attributes 'Name' and 'Age' will be returned as answer after applying query function which indicates that only these two attributes are important for generating final report to that particular query.
\par
For general case, if there exist a dataset with $n$ rows and $k$ attributes. Once the whole data is collected, the query function will be applied to the data and the attributes which is required to generate answer is returned. 
If $m$ number of attributes are important for generating the final report, then these $m$ attributes will tie up together to behave like a single attribute and the reduced number of attributes will be $g=k-m+1$. Now if $g$ is divisible by $S$, these $g$ numbers of attributes is divided into $S$ group with $g/S$ attributes in each groups. If $g$ is not divisible by $S$, then extra $e$ elements will choose a group randomly without replacement\cite{b9}. Now the whole data is divided into $t$ batches where $ith$ batch contains $n_i$  number of rows and $n_i \simeq n_j; \forall \: 1 \leq i,j \leq t$ i.e number of rows in each groups are almost equal. 
\par
Now for every batch, each group of attributes will choose a shuffler from $S$ number of shuffler randomly without replacement \cite{b9} and go for shuffling. After shuffling of first batch, the 2nd batch along with first batch where all $n_1 + n_2 \simeq 2 n_1$ number of rows will go for shuffling.  This thing will repeat every time till the last batch where the last batch will go for shuffling with total $n_1 + n_2 + ... + n_t = t n_1$ rows. Every time each batch will go for shuffling along with previous all batches and for this it is called \textit{Cumulative Iterative Shuffling(CIS)}. 
\par
As a result, the author has found that it is not necessary to apply this technique as it gives negative value of epsilon. So, here the whole concentration is to apply another approach, where the first batch with $n_1$ number of rows will go for shuffling first, then the 2nd batch with $n_2$ number of rows will go and so on. At the end, the last batch with $n_t$ number of rows will go for shuffling. This is called \textit{Iterative Shuffling(IS)}.
\par
\begin{figure}[htbp]
\centerline{\includegraphics[width = \columnwidth]{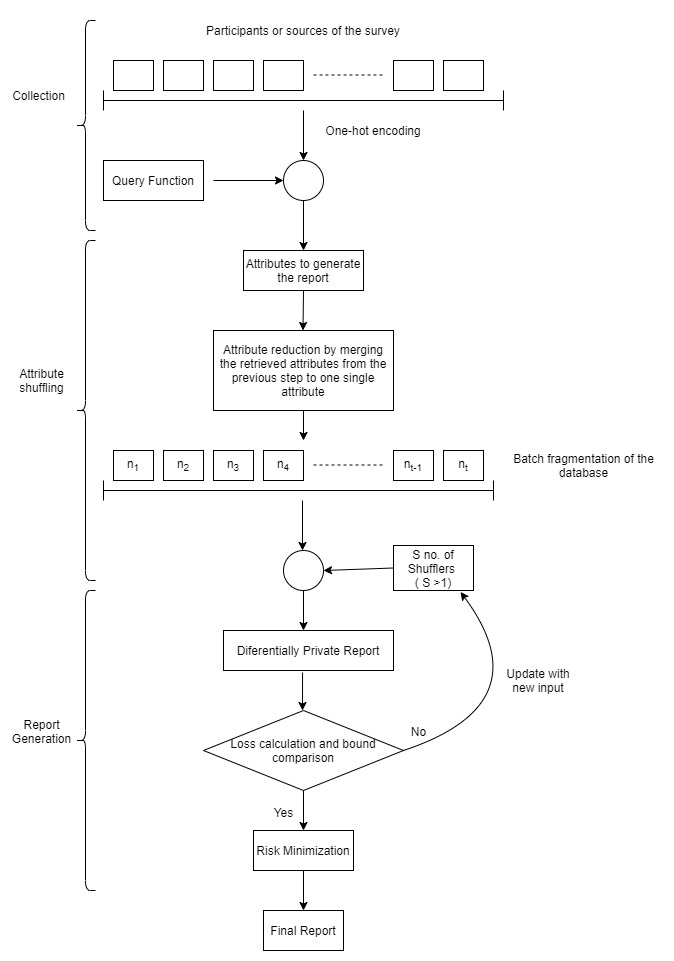}}
\caption{Architecture of BUDS}
\label{architecture_buds}
\end{figure}

Every Shuffler has separate channel for separate attributes and each shuffler has their own shuffle technique for doing independent shuffling. The whole architecture is given in the figure \ref{architecture_buds}.
\section{Privacy}\label{privacy}
In this section, the privacy budget for \textit{CIS} and \textit{IS} are derived and compared to each other. At the last part of this section, the approach with better privacy budget is found and the rest of the work is proceeded with this particular value of privacy budget.
\par
The database here is with total $n$ rows and $g$ batches and the whole database is divided into $t$ batches. Now a randomisation function is applied through shuffling, denoted by $\mathcal{R_S}$, to the data  to create strong privacy protection. 
The proof will be proceeded with the first approach ($CIS$) and the privacy budget of it will be derived here.
\newtheorem{theorem}{Theorem}
\begin{theorem}{(Cumulative Iterative Shuffling : CIS)}
A randomisation function $\mathcal{R_S}$ applied by $S\ (S > 1)$ number of shuffler providing cumulative iterative shuffling to a data set $X$ with $n$ rows and $g$ attributes, where the data base is divided into $1,2, ..., t$ batches containing $n_1,n_2, ..., n_t$ number of rows respectively, will provide $\epsilon$-differential privacy to the data with privacy budget-
\begin{equation}
    \epsilon = \ln{\bigg(\frac{1}{n_1 - 1}\bigg)^S}
\end{equation}{}
only when, $n_1 \simeq n_2 \simeq ..... \simeq n_t$.
\end{theorem}

\begin{proof}
Let, the data set contain $n$ rows and $g$ attributes and there exists $S$ number of shufflers. The attributes are divided into $S$ groups as described in the previous subsection. Each row of data set contains the information of individuals corresponding their unique ID. The data set is divided into $1, 2, ..., t$ batches containing $n_1, n_2, ..., n_t$ number of rows respectively where $n_1 \simeq n_2 \simeq ..... \simeq n_t$.\\
The first group of attributes of the first batch choose a shuffler randomly with probability $\frac{1}{S}$, The second group of attributes of the first batch choose a shuffler randomly with probability $\frac{1}{S-1}$ and so on. In the similar way the last group of attributes of the first batch will choose a shuffler only in 1 way.\\
After shuffling, a particular row will belong to their own unique Id i.e a row will be in its own position and other row will shuffle in $(n_1 - 1)!$ ways. All rows  will shuffle in $n_1 !$ ways.\\
The probability that a particular row of first batch will belong to its own unique ID and will be same as it was before the shuffling was 
\begin{align}
     \frac{1}{S} \times \frac{(n_1 -1)!}{n_1 !} \times & \frac{1}{S-1} \times \frac{(n_1 -1)!}{n_1 !} \times ... \times 1 \times \frac{(n_1 -1)!}{n_1 !}\\
     & = \frac{1}{S!} \times \bigg[\frac{(n_1 -1)!}{n_1 !}\bigg]^S
\end{align}
 The probability that the row will not belong to its own ID  after shuffling is \begin{align}
     \frac{1}{S!} \times \bigg[\frac{n_1 - (n_1 - 1)!}{n_1!}\bigg]^S
 \end{align}
The Randomised Response Ratio($RR_1$) or the probability ratio for first batch of data will be 
\begin{align}
RR_1 &= \frac{P(Row \ belongs \ to \ its \ own \ unique \ ID)}{P(Row \ does \ not \ belongs \ to \ its \ own \ unique \ ID)}\\
   &= \frac{\frac{1}{S!} \times \bigg[\frac{(n_1 -1)!}{n_1 !}\bigg]^S}{\frac{1}{S!} \times \bigg[\frac{n_1! - (n_1 - 1)!}{n_1!}\bigg]^S}\\
   &= \frac{1}{(n_1 - 1)^S}
\end{align}
The second batch will go for shuffling along with previous batch containing total $n_1 + n_2$ number of rows and here the probability ratio will be 
\begin{align}
    RR_2 = \frac{1}{(n_1 + n_2 - 1)^S}
\end{align}
Similarly the probability ratio for third batch along with all previous batches will be: 
\begin{align}
    RR_3 = \frac{1}{(n_1 + n_2 + n_3 - 1)^S}
\end{align}
This process will repeat for t batches and the probability ratio for the last batch will be: 
\begin{align}
    RR_t = \frac{1}{(n_1 + n_2 + n_3 +.....+ n_t - 1)^S}
\end{align}
The total probability ratio for the whole data base will be:
\begin{align}
    RR_\infty = &\frac{1}{(n_1 - 1)^S} + \frac{1}{(n_1 + n_2 - 1)^S} +\\ &... + \frac{1}{(n_1 +n_2 +...+n_t - 1)^S}
\end{align}
For, $n_1 \simeq n_2 \simeq ..... \simeq n_t$ and for this we can write 
\begin{equation}
    RR_\infty =  \frac{1}{(n_1 -1)^S}+ \frac{1}{(2 n_1 -1)^S}+...+\frac{1}{(t n_1 -1)^S}
\end{equation}
Now, we can see the $2^\text{nd}, 3^\text{rd},... t^\text{th}$ terms in RHS of the equation are the very small fractions and we can easily ignore it.\\
By ignoring the small terms in the equation we get,
\begin{equation}
    RR_\infty =  \frac{1}{(n_1 -1)^S}
\end{equation}
Therefore, the privacy budget $\epsilon$ will be :
\begin{equation}
    \epsilon = \ln{(RR_\infty)} = \ln{\bigg[\frac{1}{(n_1 -1)^S}\bigg]}
\end{equation}

\end{proof}
The next theorem will provide the value of the privacy budget of another approach using $IS$. 
\begin{theorem}{(Iterative Shuffling : IS)}
A randomisation function $\mathcal{R^*_S}$ applied by $S\ (S > 1)$ number of shuffler providing iterative shuffling to a data set $X$ with $n$ rows and $g$ attributes, where the data base is divided into $1,2, ..., t$ batches containing $n_1,n_2, ..., n_t$ number of rows respectively, will provide $\epsilon$-differential privacy to the data with privacy budget-
\begin{equation}
    \epsilon = \ln{\bigg(\frac{t}{n_1 - 1}\bigg)^S}
\end{equation}{}
only when, $n_1 \simeq n_2 \simeq ..... \simeq n_t$.
\end{theorem}

\begin{proof}
The first batch of the data will go for shuffling exactly as the same of the previous approach that we have already discussed earlier. So, the probability ratio for the first batch will be same.\\
The Randomised Response Ratio($RR_1$) or the probability ratio for first batch of data will be 
\begin{align}
RR_1 &= \frac{P(Row \ belongs \ to \ its \ own \ unique \ ID)}{P(Row \ does \ not \ belongs \ to its \ own \ unique \ ID)}\\
   &= \frac{1}{(n_1 - 1)^S}
\end{align}
The second batch with $n_2$ rows will go for shuffling and here the probability ratio will be 
\begin{align}
    RR_1 = \frac{1}{(n_2 - 1)^S}
\end{align}
This process will repeat for t batches and the total probability ratio for the whole data base will be:
\begin{align}
    RR_\infty = &\frac{1}{(n_1 - 1)^S} + \frac{1}{(n_2 - 1)^S} + ..... + \frac{1}{(n_t - 1)^S}
\end{align}
For, $n_1 \simeq n_2 \simeq ..... \simeq n_t$ and for this we can write 
\begin{equation}
    RR_\infty =  \frac{t}{(n_1 -1)^S}
\end{equation}
Therefore, the privacy budget $\epsilon$ will be :
\begin{equation}
    \epsilon = \ln{(RR_\infty)} = \ln{\bigg[\frac{t}{(n_1 -1)^S}\bigg]}
\end{equation}
\end{proof}
Now, the randomised response probability ratio $RR\infty$ of $CIS$ is giving a very small fraction which makes the value of $\epsilon$  negative and it is not desirable at all. But, in the next part of this section a better privacy budget can be found by using normal $IS$. This value of $\epsilon$ is giving enough privacy guarantee to continue our work. 
The work will be proceeded with $IS$ and the value of $\epsilon$ of $IS$ will be taken as the privacy guarantee for the rest of this work.\\

\section{Utility}\label{utility}
Utility can be described as how much real information we can gain from the data by a particular query or set of queries. In this section a discussion on the utility that can be gained from the data by applying BUDS is done thoroughly and a tight bound for loss function of input and output count that have a great impact on utility is provided depending on the privacy budget. At the end of this section, the authors have tried to find the optimized randomization function by minimizing empirical risk function to achieve maximum utility keeping $\epsilon$ privacy guarantee. 
\par
We have the dataset containing $n$ number of rows and $k$ number of attributes. After applying the query function, we get a data set with reduced set of attributes $g$. The whole data set is divided in $1, 2, ..., t$ batches. 
\par
Let assume, the query is \textit{"How many people are engaged with event $E$  in a time horizon $[d]$?"} where, $E \ \epsilon \ \mathcal{E}$,  $\mathcal{E} = \{E_1, E_2, ...\}$ a set of all possible events and $[d] = \{ 1, 2, ...., d\}$. Now, at the time point $T$, let the input database is $X$ and the output database is $Y$. For database $X$, that means before shuffling of the data, if a count on the people engaged with event $E \epsilon \mathcal{E}$ is taken, it will give : $c = \Sigma_\text{i $\epsilon$ [t]} \Sigma_\text{T $\epsilon$ [d]} X_i [T]$, where $i$ denotes the batch number. $\Sigma_\text{T $\epsilon$ [d]} X_i [T]$ denotes the number of people engaged with event $E \epsilon \mathcal{E}$ at the time horizon  $[d]$ in ith batch. The count $c$ gives the average number of people engaged to the event $E \epsilon \mathcal{E}$ for the whole input database at the time horizon $[d]$ which is denoted as to be actual answer of that particular query. Now, if the same count is taken on the output database. it will give : $c' = \Sigma_\text{i $\epsilon$ [t]} \Sigma_\text{T $\epsilon$ [d]} Y_i [T]$ which is the reported answer of this mechanism.
Now the main idea is to find the distance between these two counts which actually gives a great impact on utility measure. Now if the distance between these two counts are really vary close, it can be said that the utility is good enough. The aim is to minimize this distance which can be measure by the loss function.
\par
 These characteristics of BUDS hold for general cases. When the query gives $m$ number of relevant attributes, these all are also applicable in this case.
For generating report for a particular query we need to concentrate only to the tied relevant attributes which contains no extra noise except the noise produced by one-hot encoding to the data. This can be called the sub-database. Now the input sub-database $D_X \subseteq X$ and the output sub-database $D_Y \subseteq Y$ will behave like the adjacent database as the information containing in rows of $D_X$ are exactly same as $D_Y$, only difference is that the rows of $D_Y$ are shuffled, i.e each individual's information does not belongs to its own unique ID. 
Remembering this fact, for two adjacent database $D_X$ and $D_Y$ if one wants to take the count of people engaged to a event $E \epsilon \mathcal{E}$, the count $c'$ of output sub-database $D_Y$ will be in the neighbour-hood of the count $c$ of the the input sub-database $D_X$. As, in this mechanism a minimum noise is added to the data measure of distance between  $c$ and $c'$ will be vary close and only differ on $e^\text{$\epsilon$}$, where, $\epsilon$ is the privacy budget of the mechanism.\\
Thus we get the following bound:
\begin{equation}
    c \leq e^\text{$\epsilon$} c'
\end{equation}
when $\epsilon$ = 0 $\implies$ $c = c'$; i.e. The utility $\mathcal{U}(X,Y)$ reaches in highest value. If we take the range of utility $[0,1]$; when $\epsilon$ = 0 then  $\mathcal{U}(X,Y) = 1$. In this work,
\begin{equation}
    c \leq e^\text{$\ln{\bigg[\frac{t}{(n_1 - 1)^S}\bigg]}$} c'
\end{equation}
By subtracting $c'$ from both sides and taking absolute value of the equation (24) we get.
\begin{equation}
    |c - c'| \leq c' \bigg|e^\text{$\ln{\bigg[\frac{t}{(n_1 - 1)^S}\bigg]}$} - 1\bigg|
\end{equation}
We define the loss function $\mathcal{L}(c,c')$ = $|c - c'|$ and get
\begin{equation}
    \mathcal{L}(c,c') \leq c' \bigg|e^\text{$\ln{\bigg[\frac{t}{(n_1 - 1)^S}\bigg]}$} - 1\bigg|
\end{equation}
The Risk function will be calculated in the following with the idea to minimize the risk for getting maximum utility. By minimizing the risk function the optimum randomisation function will guarantee the strong privacy with maximum utility. Obviously here a hypothesis class exists that is a space containing all possible Randomisation function we are searching for. By the concept of decision theory, the main idea is to minimizing the risk function to find the best randomization function that will help us to map best from input to output. The Risk function is denoted by $Risk(\mathcal{R(S)})$ and 
\begin{equation}
    Risk(\mathcal{R(S)}) = E[\mathcal{L}(X,Y)] = \int \int P(c,c') \mathcal{L}(c,c') dc dc'
\end{equation}
Here , $P(c,c')$ is the distribution of sample data set containing $n$ data points that are drawn randomly from a population follows the distribution $\mu(Z)$ over $Z = c . c' : (c_1 , c'_1), (c_2 , c'_2), ...., (c_n , c'_n)$ and 
\begin{equation}
    P(c,c') = P(c'|c) . P(c)
\end{equation}

Now the empirical risk will be, 
\begin{equation}
    Risk_\text{(emp)} (\mathcal{R(S)}) = \frac{1}{n} \Sigma_{1}^{n} L(c_i,c'_i) \leq  \frac{1}{n} \Sigma_{1}^{n} e^\text{$\epsilon$} c'_i 
\end{equation}
We will add a regularization parameter G that is included in order to impose the complexity penalty on the loss function and prevent over fitting in the following way:
\begin{equation}
     Risk_\text{(emp)} (\mathcal{R(S)}) = \frac{1}{n} \Sigma_{1}^{n} L(c_i,c'_i) + \lambda G(\mathcal{R(S)})
\end{equation}
\begin{equation}
     Risk_\text{(emp)} (\mathcal{R(S)}) \leq  \frac{1}{n} \Sigma_{1}^{n} e^\text{$\epsilon$} c'_i + \lambda G(\mathcal{R(S)})
\end{equation}
where $\lambda$ controls the strength of complexity penalty. Let assume, we have found the mechanism $\mathcal{R*(S)}$ that minimizes the risk, then
\newcommand{\argmin}{\operatornamewithlimits{argmin}}

\begin{equation}
    \mathcal{R*(S)} = \argmin_\text{$\mathcal{R(S)} \epsilon \mathcal{H}$} Risk_\text{(emp)} (\mathcal{R(S)})
\end{equation}
Table \ref{symbol} will give the description of all the symbols that have been used in this whole discussion.

\begin{table}[htbp]
\caption{Descriptions of Symbols used}
\begin{center}
\begin{tabular}{|p{1.5cm}|p{6cm}|}
\hline
\textbf{Symbols}&\textbf{Description} \\
\hline
$n$ & Number of rows in the data set\\
\hline
$K$ & Actual number of attributes in the data set\\
\hline
$Q$ & Query function\\
\hline
$m$& Number attributes relevant the query\\
\hline
$g$ & Number of attributes when all relevant attributes tied up together.\\
\hline
$S$ & Number of shuffler\\
\hline
$t$ & The number of batches\\
\hline 
$n_i$ & Number of rows in ith batch; $i \epsilon [t] = \{1,2,...,t\}$\\
\hline
$\epsilon$& Privacy budget\\
\hline
$X$& Input database\\
\hline
$Y$& Output database\\
\hline
$D_X$& Input sub-database containing only the tied up attributes presenting a single attribute relevant to the query of the client.\\
\hline
$D_Y$& Output sub-database containing only the tied up attributes presenting a single attribute relevant to the query of the client.\\
\hline
$c$& Count related to query on the input data\\
\hline
$c'$ & Count related to query on the output data\\
\hline
$E$& A particular event\\
\hline
$\mathcal{E}$& A set of all possible events\\
\hline
$T$ & A single time point\\
\hline
$[d]$& A time horizon, $[d] = \{ 1, 2, ..., d\}$\\
\hline
$D$& Domain of the input database $X$\\
\hline

\end{tabular}
\label{symbol}
\end{center}
\end{table}
\section{Result and Discussion}
 A small example using $IS$ with $\epsilon$-DP without violating the loss function of input-output count, is given in this section. Also an empirical result is discussed later.
\par
In BUDS, a query function is used on the encoded database containing individuals information corresponding their own unique ID in each row before shuffling and it gives us the relevant attributes name to that query so that in the next step, these attributes can tie up together to present a single attribute.
\par
\textbf{Example 2: }Let's take the dataset containing names, ages, heights and weights of  six individuals where $name, age, height$ and $weight$ are four attributes and the query is \textit{"How many people are in database are less than $40 \ years$ old and have weight more than $60 \ kg$?"}. The query function returns the attributes $"weight", "Age"$ as the relevant attribute for generating final report. Here these two attributes will tie up. After that, the groups of attributes go to the shufflers for secret $IS$ where only the rows of each attributes of each batch will shuffle in a secret technique. For this, the relevant attribute pair(Weight:Age) always hold the person's weight with their own age in each row while each weight:age changes their own unique ID due to $IS$. On the other hand, rows of other irrelevant attributes ($'Name', 'Height'$) not only change their unique ID but also change their belonging weights:age due to $IS$ in a way that  one can never go back to the actual database using generated report and explore all the information related to a particular individual which is stored in the database. So, BUDS keeps a strong privacy as it is proved in section IV. The table \ref{example} is showing the data base before and after shuffling.










\begin{table}[htbp]
    \centering
    \caption{Empirical Scenario of Example 2 Before and After Shuffling}
    \begin{tabular}{|p{0.75 cm} p{0.55 cm} p{1.6 cm}|p{0.75 cm} p{0.45 cm} p{1.6 cm}|}
    \hline
         &\textbf{Before} & \textbf{IS}&&\textbf{After}&\textbf{IS} \\
         \hline
         \textbf{Name}&\textbf{Age}&\textbf{Height:Weight}&\textbf{Name}&\textbf{Age}&\textbf{Height:Weight}\\
         \hline
         Riya&20&5.3":48&Priya&7&5.3":48\\
         Sonal&7&4.8":42&Riya&20&4.8":42\\
        Priya&28&5.3":78&Sonal&28&5.3":78\\
        Sayan&35&6.00":85&Pranab&17&6.00":85\\
         Pranab&60&5.9":55&Sayan&60&6.01":64\\
         Ravi&17&6.01":64&Ravi&35&5.9":55\\
         \hline
    \end{tabular}
    \label{example}
\end{table}

\par
In BUDS, for a given query $Q$, the risk function is calculated after generating the DP report and by minimising the risk, the best randomisation function is found to generate final report. If the bound of loss function is violated for any DP report, the model will take a fresh input from client again and update the model to continue the process. In this way, the maximum utility from the data will be gained while keeping strong privacy guarantee. Table \ref{empirical} will give an empirical insight about BUDS while varying the batch size, database size, shuffler size etc. 
\begin{table}[htbp]
\caption{Empirical analytical results of the BUDS}
    \begin{center}
    \begin{tabular}{|c|c|c|c|c|}
    \hline
        \textbf{No. of rows} & \textbf{No. of batches} & \textbf{Avg. batch size}& \textbf{Shuffler} & $\epsilon$\\
         \hline
       1000   &130  &7 &3 &0.50\\
         \hline
         11000 &1000  &11 &3 & 0\\
         \hline
        100000  &5500  &18 &3 &0.11\\
         \hline
         1000000 &31000  &32 &3 &0.03\\
         \hline
         100000000 &1000000  &99 &3 &0.03\\
         \hline
         1000 &100 &10 &2 &0.2 \\
         \hline
          11000 &500 &22 &2 &0.12 \\
         \hline
          100000 &2200 &45 &2 &0.1 \\
         \hline
          1000000 &10000 &100 &2 &0.02 \\
         \hline
          100000000 &218000 &458 &2 &0.04 \\
         \hline
    \end{tabular}
    
    \label{empirical}
     \end{center}
\end{table}
\section{Future Scope and Conclusion}
In BUDS, one-hot encoding is used which transform the categorical or real valued variable in a form that helps a machine learning algorithm to produce a better result by keeping privacy with great utility. Although one-hot encoding has proved to be a good encoding option, in future other encoding options will be explored too, to further balance the trade-offs between privacy and utility. In future extensive experiments on different benchmark datasets using this algorithm has a great promise itself.
\par To summarise this whole paper, BUDS gives a balanced approach using one-hot encoding, iterative shuffling with loss management and risk minimization update. The empirical result analysis is given in Table \ref{empirical}. We get the privacy value $\epsilon = 0.02$ for the optimal conditions if we analyse the Table \ref{empirical} carefully. These results themselves are very promising for further exploration in the practical applications, as in the future scope mentioned above.  
\section{Acknowledgement}
This research was partially supported by Department of Science and Technology Grant. 
\\Grant ID: NRDMS/UG/S.Mishra/Odisha/E-01/2018.

\end{document}